\documentclass[10pt,twocolumn,letterpaper]{article}

\usepackage{iccv}
\usepackage{times}
\usepackage{epsfig}
\usepackage{graphicx}
\usepackage{amsmath}
\usepackage{amssymb}
\usepackage{algorithm}
\usepackage{algorithmic}
\usepackage{color}

 \usepackage[breaklinks=true,colorlinks,bookmarks=false]{hyperref}

 \iccvfinalcopy 

\begin{document}

\newcommand{\rb}[1]{\textcolor{red}{[Ronen: #1]}}
\newcommand{\no}[1]{\textcolor{blue}{[Nati: #1] }}

\title{Fast Detection of Curved Edges at Low SNR}

\author{Nati Ofir~~~~Meirav Galun~~~~Boaz Nadler~~~~Ronen Basri\\
Weizmann Institute\\
Dept. of Computer Science and Applied Math\\
The Weizmann Institute of Science, ISRAEL\\
}

\maketitle

\begin{abstract}
Detecting edges is a fundamental problem in computer vision with many applications, some involving very noisy images. While most edge detection methods are fast, they perform well only on relatively clean images. Indeed, edges in such images can be reliably detected using only local filters. Detecting faint edges under high levels of noise cannot be done locally at the individual pixel level, and requires more sophisticated global processing. Unfortunately, existing methods that achieve this goal are quite slow. In this paper we develop a novel multiscale method to detect curved edges in noisy images. While our algorithm searches for edges over a huge set of candidate curves, it does so in a practical runtime, nearly linear in the total number of image pixels.
As we demonstrate experimentally, our algorithm is orders of magnitude faster than previous methods designed to deal with high noise levels. Nevertheless, it obtains comparable, if not better, edge detection quality on a variety of challenging noisy images.
\end{abstract}

\section{Introduction}

\begin{figure}
\begin{center}
\fbox{\includegraphics[width=2.2cm]{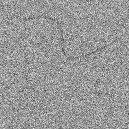}}~
\fbox{\includegraphics[width=2.2cm]{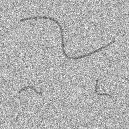}}~
\fbox{\includegraphics[width=2.2cm]{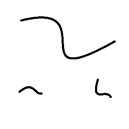}}~
\end{center}
\caption{
A noisy binary pattern with three curved fibers at high (left, SNR=1) or low (middle, SNR=2) levels of noise. The clean pattern is shown on the right. The long fiber is noticeable already at the low SNR. In contrast, the two shorter fibers can be spotted only at the higher SNR.}
\label{fig:length}
\end{figure}

This paper considers the problem of detecting faint edges in noisy images. Our key contribution is the introduction of a new, computationally efficient algorithm to detect faint curved edges of arbitrary shapes and lengths, under low signal-to-noise ratios (SNRs). A key feature of our algorithm is that the longer an edge is, the fainter it can be and still be detected. This is consistent with the performance of the human visual system, as can be appreciated from Fig.~\ref{fig:length}.

\begin{figure}
\begin{center}
\fbox{\includegraphics[height=4.8cm]{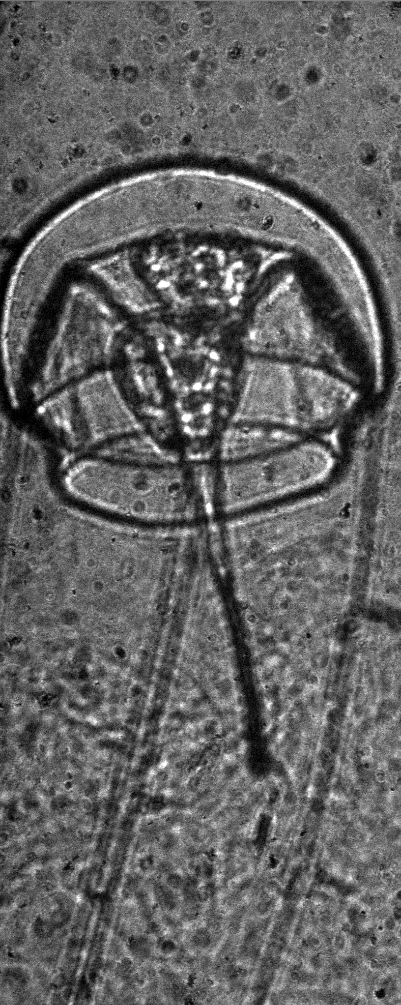}}~
\fbox{\includegraphics[height=4.8cm]{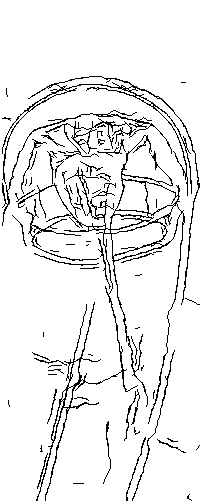}}~
\fbox{\includegraphics[height=4.8cm]{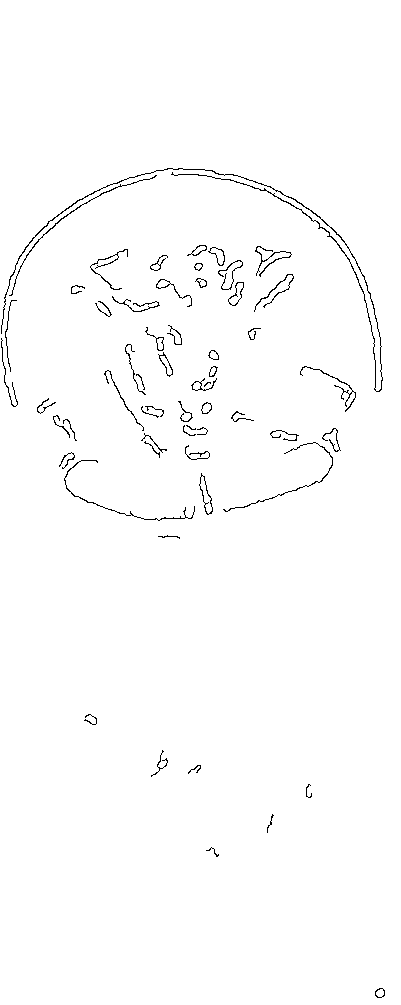}}~
\fbox{\includegraphics[height=4.8cm]{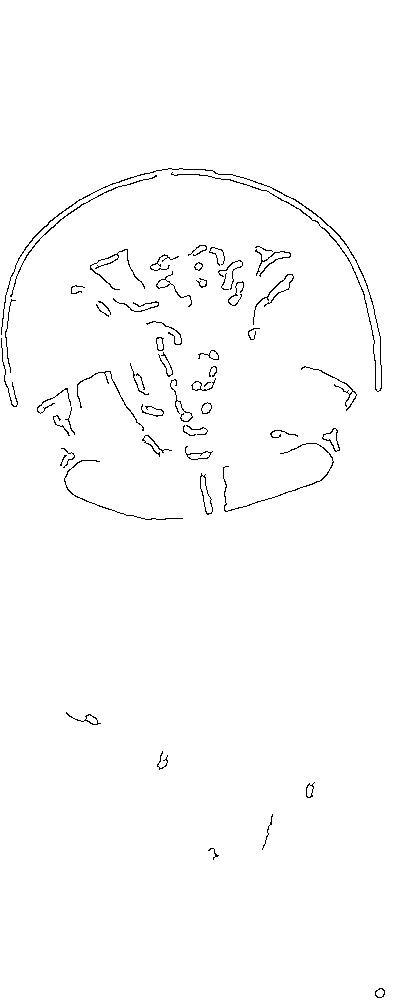}}
\end{center}
\caption{From left to right: Noisy image of a Plankton acquired by an underwater imaging system~\cite{Plankton} containing several faint edges; our result; Canny result, and  Canny after BM3D denoising. The thresholds we chose for Canny are those that maximize its performance in our experiments.} 
\label{fig:examples}
\end{figure}

Edges with low signal-to-noise-ratio are common in a variety of imaging domains, specifically when images are captured under poor visibility conditions. Examples include biomedical, satellite and high shutter speed imaging. Fig.~\ref{fig:examples} shows an image of Plankton acquired with an underwater imaging system~\cite{Plankton}. Exploring ocean ecosystems can benefit from algorithms for segmentation, detection and classification of plankton. Segmenting plankton in such images is challenging due to their low contrast and background noise. 

Noise poses an obstacle to edge detection since it can reduce the contrast of actual edges, and introduce spurious high contrasts at background pixels. As a result, common edge detection methods that are based on local image gradients are severely affected by noise. To be more robust to noise, Canny~\cite{Canny86}, for example, first denoises the image with a Gaussian filter, before computing its gradients. This smoothing operation indeed attenuates the noise, but also blurs existing edges, thus decreasing their contrast. Applying more sophisticated image denoising algorithms before edge detection does not, in general, solve the problem. 
We illustrate this point in Fig.~\ref{fig:examples}. The edges found by Canny~\cite{Canny86} on either the noisy image or the image denoised by the state-of-the art BM3D algorithm~\cite{bm3d} are shown in the two rightmost panels. Denoising prior to edge detection does not significantly improve the results. In contrast, the middle-left panel shows the edges detected by our method. 

Our approach is to detect faint curved edges by applying a large collection of \textit{matched filters} to the image, and retaining only those which are statistically significant. By definition, each of these matched filters computes the average contrast of the image pixel intensities along its corresponding curve. When this curve traces an actual edge, the filter smoothes the noise on either side of the edge while maintaining its contrast. Hence, this operation does not suffer from the degradation of the general purpose denoising approaches described above. The longer the edge and its corresponding matched filter, the more aggressively the noise is attenuated, allowing detection of fainter edges. 

The computational challenge in applying this approach is that the locations and shapes of the edges, and hence the corresponding matched filters are a-priori unknown.
Moveover, there are in general an exponential number of candidate curves and corresponding matched filters. Hence, a naive direct approach to compute all of them would be prohibitively slow. The main contribution of this paper is the development of a highly efficient algorithm to detect curved edge. This is achieved by dynamic programming on a hierarchical binary partition of the image. With this construction, even though our algorithm searches for edges over a huge set of candidate curves, it does so in a practical runtime, \textit{nearly linear} in the total number of image pixels.
As we prove in Sec.~\ref{sec:complexity}, our algorithm is orders of magnitude faster than other edge detection methods that can deal with high noise levels. Yet, as we demonstrate in Sec.~\ref{sec:experiments}, it obtains comparable, if not better, edge detection quality on challenging noisy images, from a variety of applications.
  
\section{Previous and Related Work}

Edge detection is a well studied problem with a rich history. Traditional methods considered step edges and relied on local gradients to detect them~\cite{Sobel,Laplacian,Canny86}. In contrast, recent methods addressed the problem of boundary detection in natural images~\cite{bsd,gPb,gPb-ucm,MCG,Dollar,Crisp}. These methods rely on supervised learning of complex boundary features that account for intensity, color and texture. Despite the high accuracy achieved by these methods on natural images, as shown experimentally in Sec.~\ref{sec:experiments}, they exhibit poor performance in noisy images with faint edges.

In this paper we focus on the problem of detecting step edges at high levels of noise. As mentioned in the introduction, this is an important task in a variety of imaging domains. One of the first proposed methods for step edge detection was the Sobel operator~\cite{Sobel}, which does so by thresholding the gradient magnitude at each pixel separately. Marr \& Hildreth~\cite{Laplacian} proposed to detect edges by identifying the zero crossings of a 2D Laplacian of a Gaussian filter applied to the image. These local approaches for edge detection are very efficient, essentially with linear-time complexity in the total number of pixels. However, due to their local nature, they are sensitive to noise and exhibit poor performance at low SNR. One exception is the algorithm suggested by Canny~\cite{Canny86}, which extends Sobel by hysteresis thresholding of the local gradient magnitudes. This post-processing operation significantly improves its robustness to noise. We thus choose the Canny algorithm as a baseline for comparison to our work.

A potentially promising approach to detect edges in noisy images is to first denoise the image and then apply some edge detection algorithm. The problem of image denoising received considerable attention in recent years, and various methods were proposed, including bilateral filtering~\cite{Bilateral}, anisotropic diffusion~\cite{Perona}, non-local means~\cite{nlm}, BM3D~\cite{bm3d} and more. These methods, however, are not optimized for edge detection and typically denoise the image based on small local patches. Hence, they may blur faint edges, making their detection even more difficult.

Another set of edge detection algorithms utilize a Wavelet-based bank of filters that vary in length and width, see a review in~\cite{Wavelets}. Given the availability of fast-wavelet transforms, these methods are also quite fast. However, a key difference from our approach is that these wavelet methods do not adapt to the shape of actual curved edges. Hence, their performance at very low SNRs is sub-optimal.

Several recent studies proposed to use matched filters to improve the detection of faint edges. Galun et al.~\cite{Galun} proposed an efficient algorithm for detecting straight edges, requiring $O(N\log N)$ operations for an input image with $N$ pixels. An even faster version for detecting long straight edges was proposed in~\cite{Horev}, with sub-linear run-time in the image size. An algorithm for detecting faint \textit{curved} edges was proposed by Alpert et al.~\cite{Alpert}. Relying on the Beamlet data structure~\cite{Beamlets}, their method applied dynamic programming to a hierarchical, quad-tree partition of the image. The time complexity of their algorithm is $O(N^{2.5})$, which translates into a non-practical runtime of several minutes on typical images. 

In our work, we detect curved edges with a significantly reduced complexity. The two key ideas are to instead construct a \textit{binary} tree partition, and perform a sophisticated processing on it. We present two variants of our algorithm, which both require a memory of $O(N\log N)$. The first, more stringent variant has time complexity $O(N^{1.5})$, whereas the second faster one incurs a slight loss of detection accuracy, but has even faster runtime at $O(N\log N)$ operations. For illustrative purposes, the run time of our latter algorithm, implemented in C++, is roughly 2 seconds on a small $129\times 129$ image and 10 seconds on a $257\times 257$ image. In contrast, the runtime of \cite{Alpert}, as reported in their paper is several minutes per image. 

\section{The Beam-Curve Binary Tree} \label{sec:algo}

In this paper we develop algorithms that efficiently examine an exponential set of possible candidate edges. This section describes in detail how we achieve this goal.

\subsection{Setup and Notations}

The input of our method is a noisy gray-level image $I$ with $N= m\times n$ pixels containing an a-priori unknown number of curved edges at unknown locations and shapes. For simplicity we consider square images with $m=n$ where $n$ is a power of 2. In developing our algorithm and its theoretical analysis, we consider an additive Gaussian noise model. Namely, the input image can by decomposed as $I = I_{clean}+I_{noise}$ where $I_{clean}$ is a noise free image, with step edges of constant contrast, and $I_{noise}(x,y)\sim N(0,\sigma^2)$. 

An \textit{edge} is a non self-intersecting curve $\gamma$, with step discontinuity in the pixel intensities of the unobserved $I_{clean}$. Its SNR is defined as the absolute difference in image intensities across the step edge, divided by the noise level. To each candidate curve $\gamma$, of total length $L$, passing through a set of pixels $P$, we associate the following response vector
\begin{equation} \label{eq:feature}
\phi(\gamma) = [R,L,C,P].
\end{equation}
The response value $R$, determined by the matched filter corresponding to the curve $\gamma$, is the difference between the sums of intensities on either sides of the curve. The variable $C=R/m(L)$ is the average contrast along the curve, where $m(L)$ is the total number of pixels of the matched filter of length $L$. Given two curves, $\gamma_1$ and $\gamma_2$ that share an endpoint, with the corresponding vectors $\phi(\gamma_i) = [R_i,L_i,C_i,P_i]$, $i=1,2$, the response vector of their \textit{concatenation} is 
\begin{equation} \label{eq:concate}
\phi(\gamma_1+\gamma_2) = [R_1+R_2,L_1+L_2,\frac{R_1+R_2}{m(L_1)+m(L_2)},P_1\cup P_2].
\end{equation}

To detect curves with statistically significant responses, we construct the \textit{beam-curve binary tree} of the noisy image $I$. 
This tree corresponds to a hierarchical partition of the image and an accompanying data structure, denoted $BC$, on which we provide more details below. Geometrically, each node of the tree corresponds to a tile in the image (a region of a prescribed shape). A tile $V$ of area $A$, is split into two sub-tiles $V_1,V_2$ of roughly equal area by an \textit{interface} (a straight line in our implementation) whose length is proportional to $\sqrt{A}$. The root node, at level $j=0$ represents the full image, whereas at each finer level $j$ there are $2^{j}$ tiles. The tree is constructed up to a bottom level $j_{b}$, whereby all of its tiles have area approximately $n_{\min}\times n_{\min}$, where $n_{\min}$ (typically a few pixels) is a user chosen parameter. 

Within each tile $V$, for every pair of pixels $p_1$ and $p_2$ on its boundary $\partial V$, the data structure $BC$ stores a response vector $\phi(\gamma)$. Postponing exact definitions to Sec.~\ref{sec:threshold}, this response vector corresponding to a single curve $\gamma$  between $p_1$ and $p_2$, which is most likely to be an edge. See Alg.~\ref{alg:main} for a pseudo-code of the tree construction, and Fig.~\ref{fig:tree} for its implementation based on a rectangular tile partition.

\subsection{The Edge Detection Algorithm}\label{sec:outline}

As described in Alg.~\ref{alg:main}, the matched filter responses of the various curves are calculated in a bottom-up fashion, from the leaves of the tree to its root. For each leaf tile $V$ at the bottom level $j_{b}$, we set the response of each pair of boundary pixels $p_1,p_2\in\partial V$ to be the matched filter response of the straight line segment that connects them. See Alg.~\ref{alg:bottom} for a pseudo-code of the bottom level processing.

Next, given the response vectors at level $j+1$ we compute responses at the next coarser level $j$. We do this by concatenating sub-curves from sibling tiles as follows. Let $V_1,V_2$ be sibling tiles at level $j+1$, with a parent tile $V$ at level $j$, and let $V_{12}=\partial V_1\cap \partial V_{2}$ be their joint interface. We consider all pairs of pixels such that $p_1\in \partial V_1\cap\partial V$ and $p_2\in\partial V_2\cap\partial V$. Now, for each such pair of pixels $(p_1,p_2),$ in the stringent variant of our algorithm, we consider all pixels in the joint interface $V_{12}$. For any such pixel $p_3 \in V_{12}$ we compute the response of the curve obtained by concatenating the two sub-curves from $p_1$ to $p_3$ and $p_3$ to $p_2$. Among all of these concatenated curves, we store the one with highest response score (defined explicitly in Eq.~\eqref{eq:score} below). See Alg.~\ref{alg:merge} for pseudo-code of the coarser level construction.

The final output of our algorithm is a soft edge map image $E$. A value $E_{ij}=0$ means that there is no edge passing through pixel $(i,j)$, whereas the higher the value of $E_{ij}$, the stronger is our belief that an edge passes there. This edge map $E$ is constructed as follows: We initially set all pixels of $E$ to zero. For every response vector $\phi(\gamma)$, as detailed in Sec.~\ref{sec:threshold}, we assign a significance score that depends on its mean contrast $C$ and its length $L$. A positive score indicates that the curve marks a statistically significant edge. Then, we sort all the curves in the tree whose score is positive from highest score to lowest. For each curve $\gamma$ in this sorted list, for each pixel $p\in P$ we set $E(p)$ to hold the score of $\gamma$. To deal with overlapping curves with positive scores, we apply the following simple non-maximal suppression procedure: First, if at some pixel $p$, the current value $E(p)$ is already positive, we do not decrease it; and, secondly, if most of the pixels in the $P$ were already marked by previous curves (of higher scores), we discard the current curve $\gamma$ and do not add its pixels to $E$.

In principle, the generic algorithm above can be implemented with any binary partition of the image. In Sec.~\ref{sec:complexity} we make a detailed complexity analysis, under the assumption that the children tiles are roughly half the area of their parent tile and the length of their interface is roughly the square root of their area. We prove that under these conditions, the algorithm complexity is $O(N^{1.5})$. 

\begin{algorithm}
\caption{$BeamCurveTree(V)$}
\label{alg:main}
\begin{algorithmic}
  \REQUIRE Tile $V$ whose maximal side length is $n$.
    \IF{$n\le n_{\min}$}
        \STATE $BC \gets BottomLevel(V)$
        \ELSE

    \STATE $V_1,V_2 \gets SubTiles(V)$
    \COMMENT{ The tile is split into two sub-tiles of equal area} 
    \STATE $BC_1 \gets BeamCurveTree(V_1)$
    \STATE $BC_2 \gets BeamCurveTree(V_2)$
    \STATE $BC \gets CoarserLevel(V,V_1,V_2,BC_1,BC_2)$
    \ENDIF
    \RETURN $BC$
\end{algorithmic}
\end{algorithm}

\begin{algorithm}
\caption{$BottomLevel(V)$}
\label{alg:bottom}
\begin{algorithmic}
  \REQUIRE Small tile $V$.
  
        \STATE $BC \gets EmptySet$
        \FOR{$\forall p_1,p_2 \in \partial V$} 
                \STATE $\gamma \gets$ straight line from $p_1$ to $p_2$
                \STATE $BC.add(\phi(\gamma))$
        \ENDFOR
        \RETURN $BC$
\end{algorithmic}
\end{algorithm}

\begin{algorithm}
\caption{$CoarserLevel(V,V_1,V_2,BC_1,BC_2)$}
\label{alg:merge}
\begin{algorithmic}
  \REQUIRE $V$ is an image tile, $V_1$ and $V_2$ are its sub-tiles.
  \REQUIRE $BC_1$ is a set of the responses of sub-tile $V_1$.
  \REQUIRE $BC_2$ is a set of the responses of sub-tile $V_2$.

  \STATE $BC \gets$ $BC_1\cup BC_2$
  \IF{BasicMode}
  \STATE $InterfaceSet \gets \partial V_1 \cap \partial V_2$
  \ELSIF{OptimizedMode}
  \STATE $InterfaceSet \gets BestPixels(\partial V_1 \cap \partial V_2)$
  \ENDIF
        \FOR{$\forall p_1,p_2: p_1\in\partial V\cap \partial V_1,p_2\in\partial V\cap \partial V_2$} 
             \STATE $AllResponses \gets EmptySet$ 
                    \FOR{$\forall p_3 \in InterfaceSet$}
                                \STATE $\gamma_1 \gets$ curve from $p_1$ to $p_3$ in set $BC_1$
                                \STATE $\gamma_2 \gets$ curve from $p_3$ to $p_2$ in set $BC_2$
                                \STATE $\phi(\gamma) \gets concatenate(\phi(\gamma_1),\phi(\gamma_2))$
                                \STATE $AllResponses.add(\phi(\gamma))$
                                \ENDFOR
                        \STATE $BC.add(AllResponses.bestResponse())$
        \ENDFOR
        \RETURN $BC$
\end{algorithmic}
\end{algorithm}

\begin{figure}[tb]
\centering
\includegraphics[width=4cm]{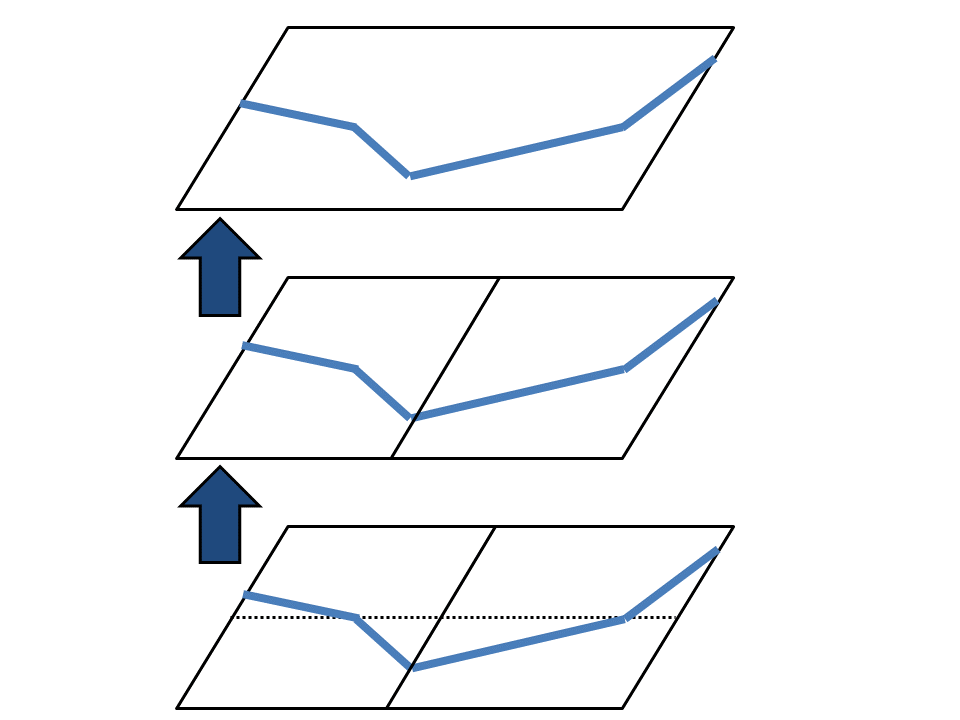}~~
\includegraphics[width=4cm]{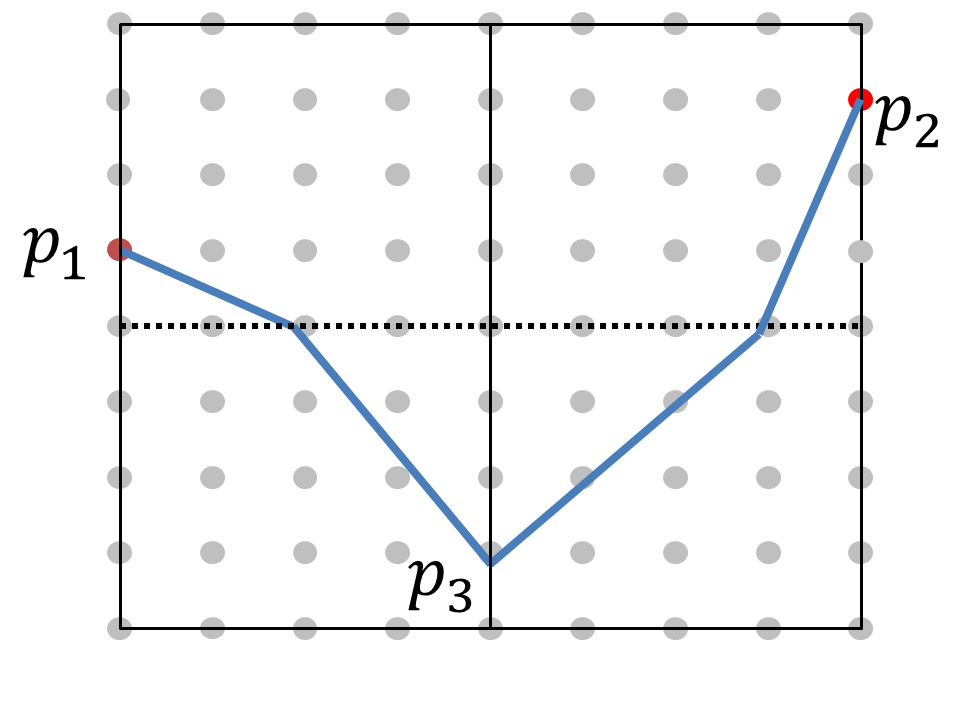}~~
\\[0.1cm]

\caption{Left: The 3 topmost levels of the Rectangle-Partition-Tree. Right: The $n\times n$ image at level $j=0$ is partitioned into two rectangles of size $n \times n/2$ at level $j=1$. Each rectangle is then partitioned into two $n/2 \times n/2$ squares at level $j=2$. 
A curve going from two boundary pixels $p_1,p_2$ of level $j=0$ is a concatenation of up to 2 sub-curves of level $j=1$, and up to 4 sub-curves of level $j=2$.}
\label{fig:tree}
\end{figure}

\vspace{0.1cm}
\textbf{Rectangular partition}. In this partition, we split the square image into two rectangles of equal size. We next split each rectangle into two squares and continue recursively until squares of a size $n_{\min}\times n_{\min}$ are obtained. Specifically in our implementation, we used $n_{\min}=5$. We call the obtained data structure \textit{Rectangle-Partition-Tree} (RPT). See Fig.~\ref{fig:tree} for an illustration of the RPT levels. 

\begin{figure}[tb]
\centering
\includegraphics[width=4cm]{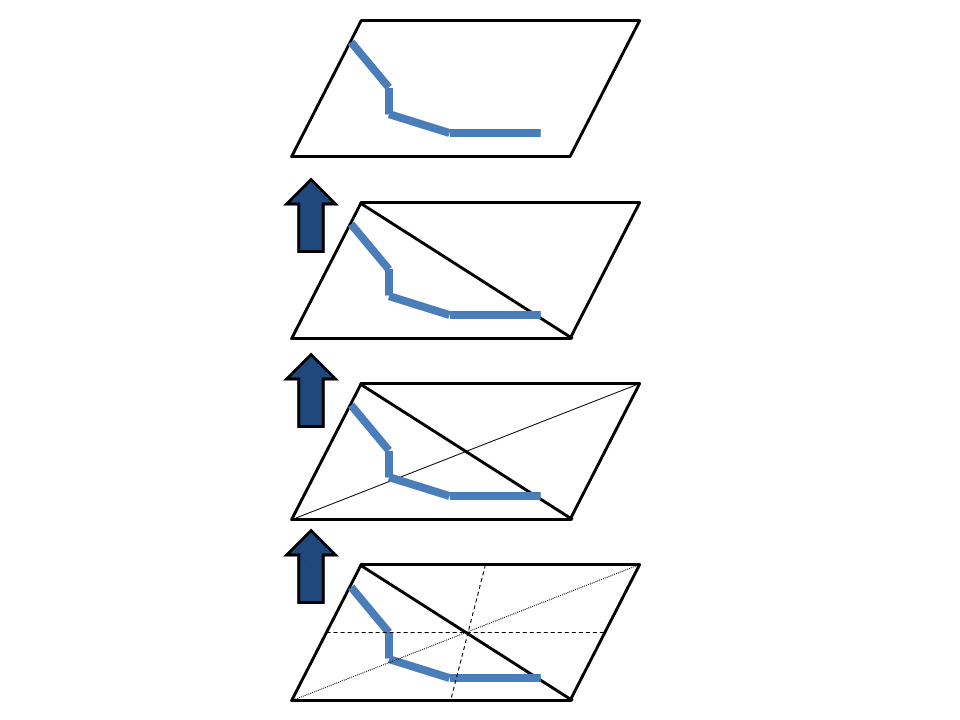}~~
\includegraphics[width=4cm]{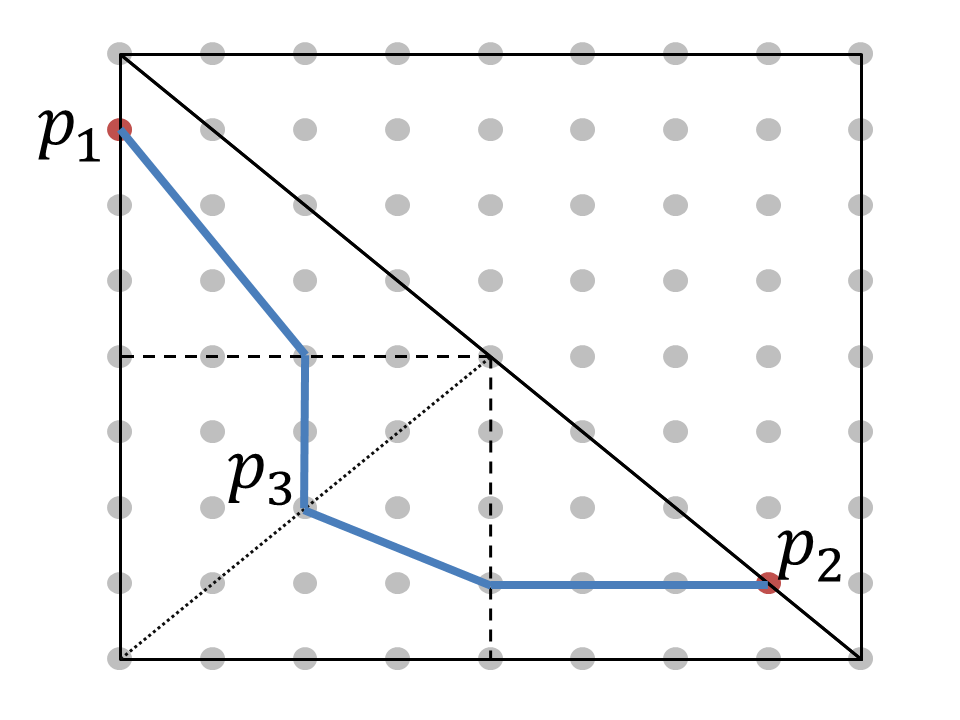}~~

\caption{Triangle-Partition-Tree, an alternative implementation of the beam-curve binary-tree. Left: The 4 topmost levels of the TPT. Right: The partitioning embedded in the 2D image grid, every triangle is divided into two sub-triangles.}
\label{fig:triangle}
\end{figure}

\vspace{0.1cm}
\textbf{Triangular Partition.}
A second possible beam curve binary tree, which we denote as the \textit{Triangle-Partition-Tree} (TPT), is based on triangular tiles, see Fig.~\ref{fig:triangle}. Here, at the topmost level we split the image along its diagonal into two triangles. Then, at any subsequent level we split each triangle into two sub-triangles recursively. For a square image, the triangles obtained are all right angled isosceles. It can be shown that the TPT-based edge detection algorithm is slightly faster than the RPT-based one. Specifically, for a square image of $N$ pixels, the TPT algorithm requires $\approx 14N^{1.5}$ operations (derivation omitted due page limit restrictions). In contrast, as we analyze in detail in Sec.~\ref{sec:complexity} below, the RPT construction requires $\approx 18N^{1.5}$ operations. In addition, the set of potential curves scanned by both partitions is of comparable size. Hence, both partitions yield a very similar edge detection performance. 

\subsection{An Optimized Version} \label{sec:optimized}

A run-time complexity of $O(N^{1.5})$ operations may still be too slow when processing large images. Hence, it is of interest to develop even faster algorithms. Below we introduce an efficient variant of our algorithm that achieves an $O(N\log N)$ complexity, at a statistical price of slightly worse detection quality.

In this variant, at each tile $V$ of level $j$ for every pair of points $p_1, p_2 \in \partial V$, the algorithm still seeks to compute the curve between them with best response. However, instead of scanning all the points in the joint interface $V_{12}=\partial V_1 \cap \partial V_2$, it only accesses a subset of $k$ pixels for some fixed constant $k$. To select this subset, for each pixel $p_{3}\in V_{12}$ we look at the curve with highest response, that starts at either $\partial V_1$ or $\partial V_2$ and ends at $p_3$, as already previously computed at level $j+1$. We then keep only the $k$ pixels with highest responses. As we prove in Sec.~\ref{sec:complexity}, the overall required number of operations of this variant is significantly smaller. Our experiments in Sec.~\ref{sec:experiments} indicate that in practice this results in a negligible decrease in edge detection quality.

\section{Complexity Analysis}\label{sec:complexity}

In this section we study the computational complexity of the beam-curve algorithm of Sec.~\ref{sec:outline}. We begin with an analysis of the general algorithm as follows. Denote by $f(A)$ the number of operations performed by our algorithm on a tile $V$ of area $A$.
Denote by $V_1$ and $V_2$ the children tiles of $V$, $V_1\cup V_2 = V$. The step of populating the responses of $V$ involves computing responses for every triplet of pixels, $p_1\in\partial V1$, $p_2\in\partial V_2$ and $p_3\in\partial V_1 \cap \partial V_2$. The length of each of these three boundaries is $O(\sqrt A)$, and so the complexity of this step is proportional to $A^{1.5}$. The operation is then repeated for the sub-tiles $V_1$ and $V_2$ whose areas $\approx A/2$. Therefore $f(A)$ satisfies the following recursion, 
\begin{equation}
f(A) = 2f(A/2)+O(A^{1.5}).
\end{equation}
Since at the bottom level of the tree, $f(n_{\min}\times n_{\min})=O(1)$, the master theorem~\cite{master} yields that $f(A) = O(A^{1.5})$. Finally, as the area of the root tile equals the total number of image pixels $N$, the overall complexity of the beam-curve binary tree algorithm is thus $O(N^{1.5})$ operations.

Next, we now explicitly bound the multiplicative constant hidden in the above $O$ notation, as well as compute the required memory for the rectangular partition. Recall that in the RPT, the root level $j=0$ represents the original $n\times n$  image, whereas the next level, $j=1$, contains two equal rectangles of size $n\times n/2$. In general, every even level $j$ includes squares of size $n/2^{j/2} \times n/2^{j/2}$ pixels, while every odd level $j$ includes rectangles of size $n/2^{(j-1)/2} \times n/2^{(j+1)/2}$. 

We next derive the number of stored responses per level. At every tile of even level $j$, we store the edge responses of curves that pass from one side of a tile to a different side. The number of stored responses thus equals the product of 4 choose 2 (pairs of rectangle sides) with the tile area,
\begin{equation}  \label{eq:curvespertile}
{4\choose 2}\cdot n/2^{j/2}\times n/2^{j/2} = 6N/2^j.
\end{equation}
This equation holds roughly also for the odd levels. Consequently, multiplying Eq.~\eqref{eq:curvespertile} by the number of tiles, gives that the total number of stored curves at level $j$ is 
\begin{equation} \label{eq:totalNumberOfCurves}
6N/2^j \times 2^j = 6N.
\end{equation}

\textbf{Space Complexity.} Since the number of curves in each level is $6N$, the number of levels is bounded by $\log N$ and the storage per curve is constant, the total required space is $O(N\log N)$.

\vspace{0.1cm}
\textbf{Basic Algorithm Complexity.} First, recall that at the bottom level, for every leaf we scan all the straight lines that begin on one side of the tile and end on another side. Since the number of operations required to compute each such response is bounded by a constant, the total number of operations at this level is $O(N)$.

Next, at every coarser level $j$, we build each curve by concatenating two shorter curves of level $j+1$, that share a common endpoint at the joint interface.
The work done at level $j$ is thus bounded by the number of calculated curves times the length of  the tile joint interface. Since the former is $6N$, the overall number of operations in an even level is 
\begin{equation}
6N \times n/2^{j/2} = 6N^{1.5}/2^{j/2},
\end{equation}
while the number of operations in an odd level is 
\begin{equation}
6N \times n/2^{(j+1)/2} = 6N^{1.5}/2^{(j+1)/2}.
\end{equation}
Denote by $C(N)$ the time complexity of the algorithm that considers all the possible points on the interface. By summing these over all scales we get the following,
\begin{equation}
C(N)\approx6N^{1.5}\left[\sum_{j\ even}^{j_b}2^{-j/2}+\sum_{j\ {odd}}^{j_b}2^{-(j+1)/2}\right].
\end{equation}
Consequently,
\begin{equation}
C(N) \leq  6N^{1.5}\left[\sum_{l=0}^{\infty}2^{-l}+\sum_{l=1}^{\infty}2^{-l}\right]=18N^{1.5}.
\end{equation}

\textbf{Optimized Algorithm Complexity.} As described in Sec.~\ref{sec:optimized}, the optimized variant scans, for each of the $6N$ pairs of start and end points, only the best $k$ pixels in the joint interface. Overall, it goes over $6Nk$ curves at each level $j$. Additional work is required to select the best $k$ pixels. Since the number of tiles equals $2^j$ and the interface length is $\approx n/2^{j/2}$, the number of operations of this step is bounded by $\log k\times 2^j\times n/2^{j/2}<N$ for every level $j$. To conclude, the total number of operations at each level is no more than $(6k+1)N.$ Thus, 
the overall complexity is bounded by $(6k+1)N\log N$, which significantly improves the overall runtime.

\section{Detection Threshold and Search Space Size} \label{sec:threshold}

As described in Sec.~\ref{sec:algo}, our method scans a huge set of candidate curves. For many pairs of start and end pixels, the algorithm stores in the data structure $BC$, the curved edge with highest edge score, as defined by Eq.~\eqref{eq:score} below. Clearly, for most images, the vast majority of these responses do not trace actual image edges and should be discarded. This task requires the determination of a suitable threshold, possibly dependent on edge length, such that only edge responses above it are retained.   

Previous work~\cite{Alpert} introduced such a theoretical threshold, designed to control the average number of false positive detections.
To this end, consider a pure noise image $I=I_{noise}$, where $I_{noise}(x,y)\sim N(0,\sigma^2)$. By definition, this image contains no real edges, and so, with high probability, all of its edge responses should be discarded. Suppose that there are $K_{L}$ distinct candidate edges of length $L.$
Then, the corresponding threshold, $T(L,K_L),$ is approximately the maximal edge contrast expected in $I_{noise}$ among $K_L$ statistically independent curves of length $L$. Alpert et al.~\cite{Alpert} showed that 
\begin{equation} \label{eq:threshold}
T(L,K_L) \approx \sigma \sqrt{\frac{2 \ln {K_L}}{wL}},
\end{equation}
where $w$ is the width of the matched filter. Thus, to each curve of length $L$, we assign an \textit{edge score}, defined as the difference between its mean contrast and the threshold,
\begin{equation} \label{eq:score}
Score(\gamma)=C(\gamma) - T(L,K_L).
\end{equation}
A positive score indicates that the candidate curve traces an edge. Moreover, higher scores represent stronger confidence for this indication.

To compute the edge score function of Eq.~\eqref{eq:score}, we thus need to know, for each curve length $L$ the size of the corresponding search space size, $K_L$. As shown below in Eq.~\eqref{eq:kl}, for the RPT construction, this quantity is approximately given by $K_{L}\approx 6N\times2^{\beta L}$ for a suitable constant $\beta$. Inserting this expression into Eq.~eq\eqref{eq:threshold} gives 
\begin{equation} \label{eq:threshold2}
T(L) = \sigma \sqrt{\frac{2 \ln({6N\times2^{\beta L})}}{wL}}.
\end{equation}

\begin{figure}
\begin{center}
\fbox{\includegraphics[width=4.5cm]{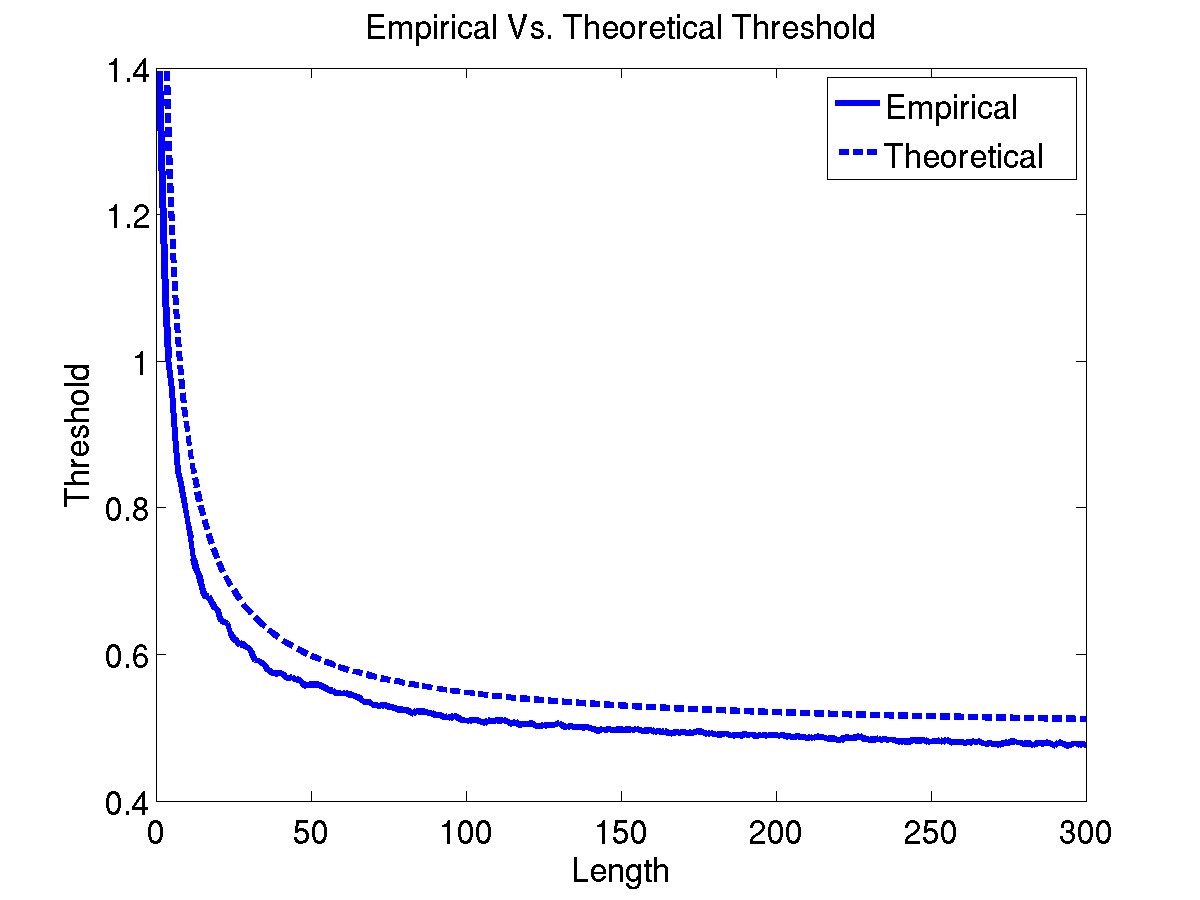}}
\end{center}
\caption{Mean contrast threshold as a function of curve length. Our threshold is designed to eliminate all the false positive detections with high probability. Theoretical curve is based on~\eqref{eq:threshold2}. Empirical threshold is produced by applying our algorithm on a pure noise images, and storing the maximal response for each curve length.}
\label{fig:threshold}
\end{figure}

\textbf{Minimal Detectable Contrast.} An interesting question, already raised by~\cite{Alpert}, is how faint can an edge be and still be detected. In our case, note that as $N$ and $L$ tend to infinity, the threshold in Eq.~\eqref{eq:threshold2} converges to a finite limit,
\begin{equation}
T_\infty = \Omega(\frac{\sigma}{\sqrt{w}}).
\end{equation}
Namely, our threshold is bounded from below by a positive constant. Hence, our ability to detect faint edges of unknown shape and location, in low SNR is limited. Fig.~\ref{fig:threshold} compares the theoretical threshold of Eq.~\eqref{eq:threshold2} to empirical results. The latter was computed by running our algorithm on a pure noise image and storing for each curve length the maximal response that was obtained. It can be seen that both curves are close one to each other, and that the graphs converge to $\approx 1/2$. This value is the asymptotic bound $T_\infty$ for the selected parameters in this simulation: width $w = 4$, image size $N = 129^2$, noise level $\sigma = 1$ and $\beta = 0.65$.

\subsection{Search Space Size}

We now compute the size of the search space $K_{L}$ of candidate curves of length $L$ in the RPT. This quantity directly affects the contrast threshold~\eqref{eq:threshold}.  

We first calculate the search space size of level $j$ of the RPT, and then we show its connection to $K_L$. Denote by $S(j)$ the total number of candidate curves at level $j$, and denote by $S'(j)$ the same number, but for given fixed start and end points. Since by Eq.~\eqref{eq:totalNumberOfCurves}, the total number of stored curves at any level is equal to $6N$, then
\begin{equation} \label{eq:S_S_tag}
S(j) = 6N\times S'(j).
\end{equation} 
Next, to compute $S'(j)$, recall that in the RPT, we split a tile $V$ of level $j$ into two sub-tiles $V_1,V_2$ of level $j+1$, by a joint interface $\partial V_1 \cap \partial V_2$, whose length is $\approx n/2^{j/2}$. For fixed endpoints $p_1,p_2\in\partial V$, the quantity $S'(j)$ satisfies the following recursive formula
\begin{equation}
S'(j) = S'(j+1)\times S'(j+1)\times n/2^{j/2}.
\end{equation}
In order to apply the master theorem~\cite{master}, we take logarithm on both sides
\begin{equation}
\log S'(j) = 2\log S'(j+1)+ \frac{1}{2}\log (N/2^{j}).
\end{equation}
Substitute $A=N/2^{j}$, which is exactly the number of pixels in a tile at level $j$, and define $\tilde S(A) = S'(j)$. Then,  
\begin{equation}
\log \tilde S(A) = 2\log \tilde S(A/2)+ \frac{1}{2}\log A.
\end{equation}
According to~\cite{master}, $\log \tilde S(A)=O(A)$. Subsequently, $S'(j) =  2^{O(N/2^{j})}$ and combining this with Eq.~\eqref{eq:S_S_tag}, 
\begin{equation} \label{eq:sp}
S(j) =  6N\times2^{O(N/2^{j})}.
\end{equation}
To derive an expression for $K_L$, it can be shown, that the average length over all candidate curves in a given tile at level $j$ is proportional to the tile's area: $L =  O(N/2^{j}).$  Therefore we can approximate $K_L$, by
\begin{equation} \label{eq:kl}
K_L \approx 6N\times2^{\beta L}
\end{equation}
for some constant $\beta$. As mentioned above, we showed empirically that $\beta = 0.65$.

We remark that even though the search space is exponential in $L$, there are nonetheless various curves that cannot be found by the RPT. Examples of such curves include closed curves like a circle, self-intersecting curves like a cross, and also any curve that begins and ends in the same side of a tile, like a narrow 'V' shape. Such geometric objects are detected by our algorithm as union of shorter valid curved edges.  

\section{Experiments}
\label{sec:experiments}

\begin{figure}
\begin{center}
\fbox{\includegraphics[height=3cm]{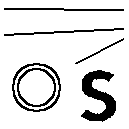}}~
\fbox{\includegraphics[height=3cm]{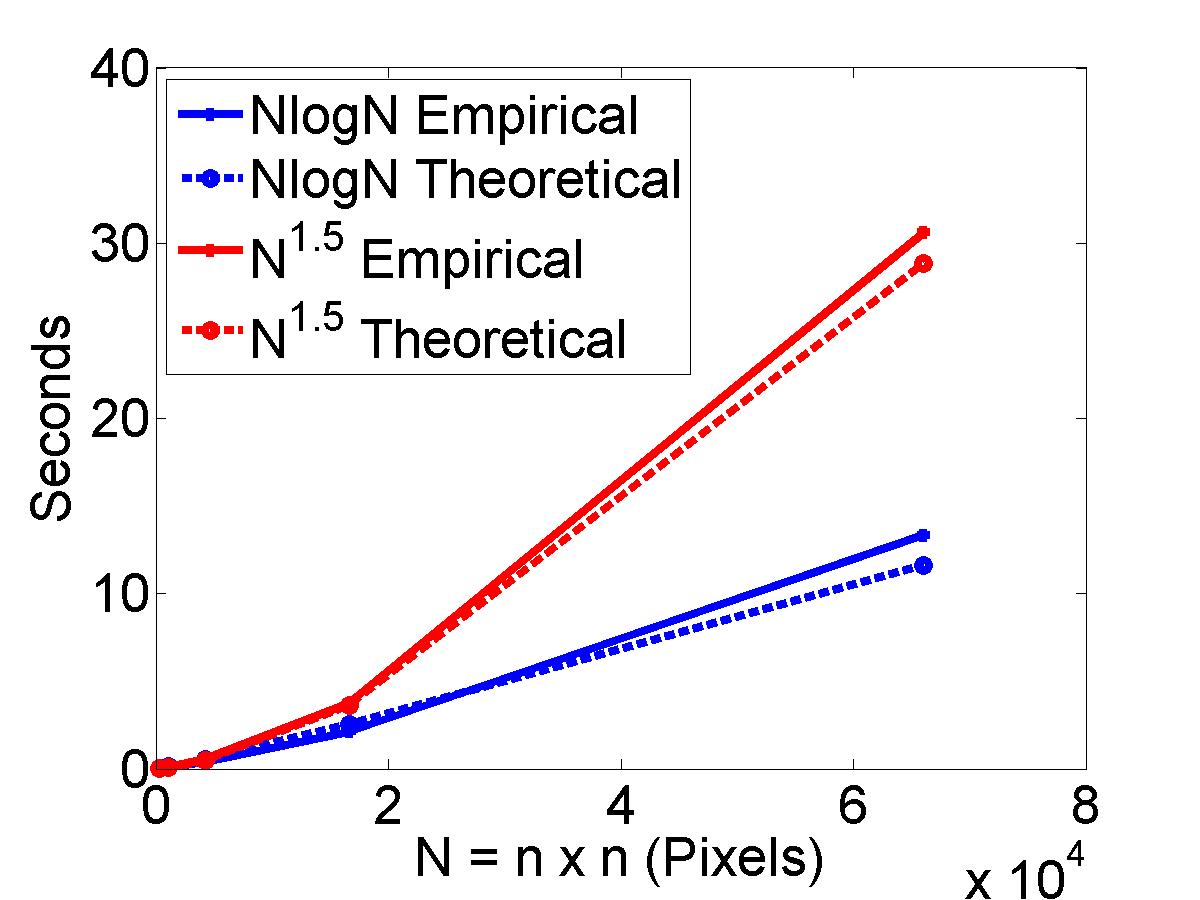}}
\end{center}
\caption{Left: A binary pattern used in simulation experiments.
Right: Empirical run-time of our implementation compared to the theoretical run-time for both $O(N\log N)$ and $O(N^{1.5})$ modes.}
\label{fig:simulation}
\end{figure}

We tested our algorithm both on simulated artificial images, and on challenging real images from several application domains. Our code is implemented in c++. Furthermore, as our tree construction can be easily parallelized, our code uses multi-threading. We ran our experiments on an 8-core Intel i7, 16 GB RAM single machine. For an input image of size $129\times 129$ pixels our run times are $\approx 2$ seconds for the $O(N\log N)$ version and $\approx 4$ seconds for the $O(N^{1.5})$, both including the post processing step of computing the final edge map image $E$. For an input image of $257\times 257$ pixels the corresponding run-times are $\approx10$ and $\approx 30$ seconds respectively. More CPU cores or utilization of a GPU can reduce these run times significantly. Fig.~\ref{fig:simulation} shows the empirical and theoretical run-times as a function of image size.
Our implementation outputs the edge image together with a list of all the curved edges that passed the statistical threshold.
We compared our solution to the classical Canny~\cite{Canny86} algorithm, and to several state of the art algorithms for boundary detection in natural images, including Multiscale Combinatorial Grouping (MCG)~\cite{MCG}, Crisp boundary detection~\cite{Crisp} and Dollar and Zitnick~\cite{Dollar}.

\vspace{0.1cm}
\textbf{Simulations.} For the simulations, we used a binary pattern of size $129\times 129$ pixels, contains straight lines, concentric circles and an 'S' shape, see Fig.~\ref{fig:simulation}. We then scaled the intensities and added additive Gaussian noise with $0.1$ standard deviation to produce images with SNRs between $0.6$ and $2.6$ in intervals of $0.2$. In addition, we add a very weak salt and pepper noise that affects only 1\% of the pixels in the images. We used the result of Canny on the clean binary pattern as the ground truth. We quantitatively compared the various edge detection algorithms by computing their Precision $(P)$, Recall $(R)$, and F-score, using the popular F-measure for image segmentation described in \cite{bsd}. 

\begin{figure}[t]
\begin{center}
\fbox{\includegraphics[width=6.0cm]{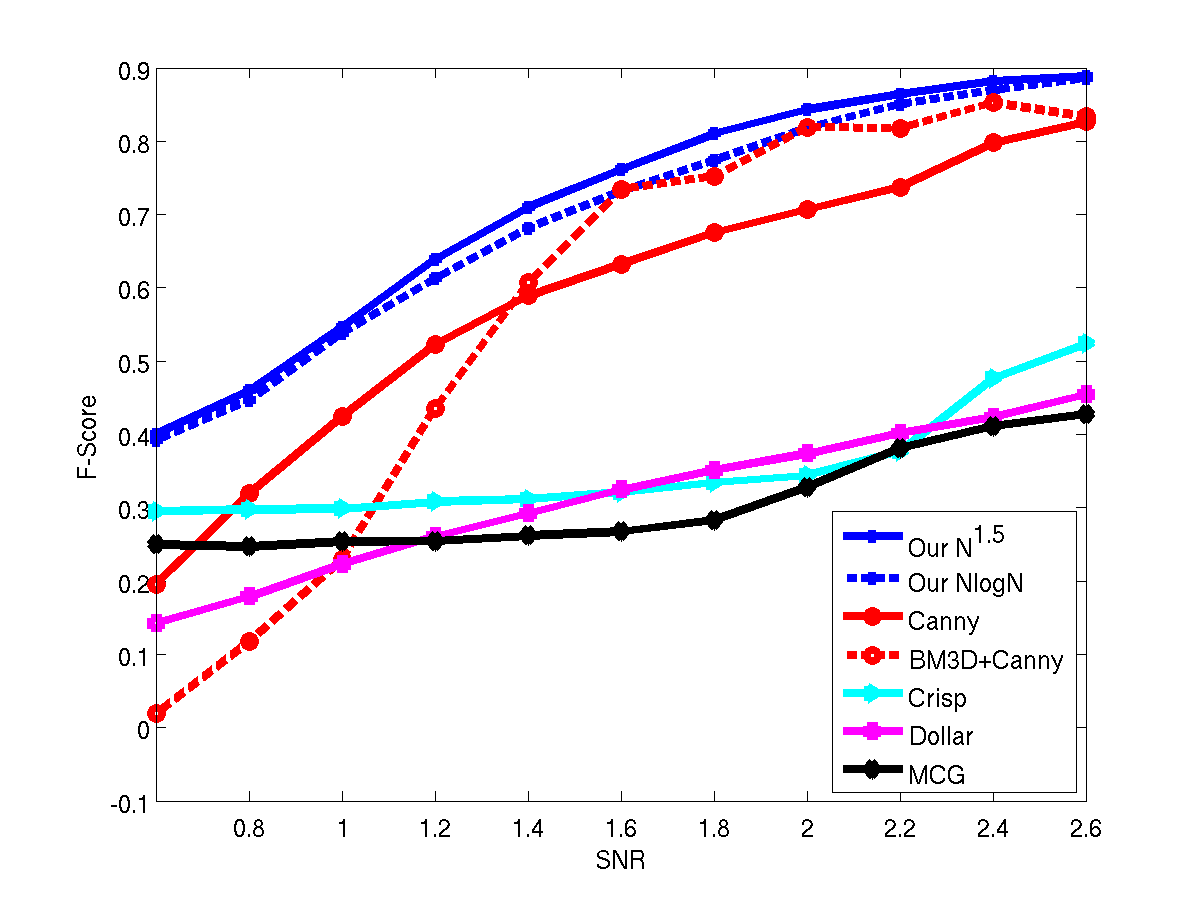}}
\end{center}
\caption{Simulation results: F-measures obtained with various edge detection algorithms as a function of SNR.}
\label{fig:F-measure}
\end{figure}

\begin{figure}\label{fig:cc}
\centering
\fbox{\includegraphics[width=1.8cm]{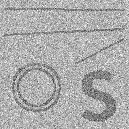}}~
\fbox{\includegraphics[width=1.8cm]{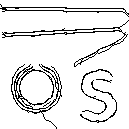}}~
\fbox{\includegraphics[width=1.8cm]{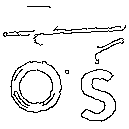}}~
\fbox{\includegraphics[width=1.8cm]{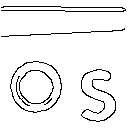}}
\caption{Result of applying various edge detection algorithms to the noisy simulation image. From left to right: Input noisy image at SNR 2, our $O(N^{1.5})$ algorithm, Canny best result and Canny after BM3D denoising.}
\label{fig:simulations}
\end{figure}

Table~\ref{tab:F-measure} shows the average F-scores obtained for each algorithm. A graph of the F-scores as a function of SNR is shown in Fig.~\ref{fig:F-measure}. As expected, algorithms geared to detecting boundaries in natural images do not perform well in such noisy conditions. Canny manages to obtain fairly high F-scores. It is important to note that to obtain these results we had to modify Canny's default thresholds. We thresholds used $[low,high] = [0.26,0.65]$ were in fact optimized over to maximize its F-score in our simulations. In contrast, it can be seen that our $O(N^{1.5})$ algorithm achieves the best F-scores at all SNR levels, while our $O(N\log N)$ variant achieved only slightly smaller scores. These results suggest that high quality detection can be obtained by our faster method, in spite of the significant reduction of complexity. Fig.~\ref{fig:simulations} shows representative results of Canny and our method at SNR=2. 

\begin{figure}[t]
\begin{center}
\fbox{\includegraphics[width=1.8cm]{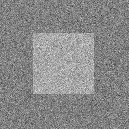}}~
\fbox{\includegraphics[width=1.8cm]{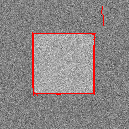}}~
\fbox{\includegraphics[width=1.8cm]{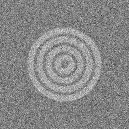}}~
\fbox{\includegraphics[width=1.8cm]{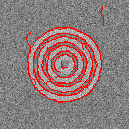}}
\end{center}
\caption{Our results on the simulations of Alpert et al.~\cite{Alpert} at SNR 2.}
\label{fig:alpert}
\end{figure}

We also made extensive evaluation of our algorithm on the dataset of $63$ artificial images containing various straight and curved edges, as studied by~\cite{Alpert}. Fig.~\ref{fig:alpert} shows some representative results of our algorithm on few of these images, shapes at different SNRs. On this dataset our $O(N^{1.5})$ algorithm achieved an average F-score of $\approx 0.79$, compared to $\approx 0.74$ achieved by~\cite{Alpert}. This demonstrates that our algorithm is not only significantly faster, but also at least as accurate as~\cite{Alpert}.

\begin{table}[h]
\begin{center}
\begin{tabular}{|l|c|c|c|}
\hline
\textbf{Algorithm $\backslash$ SNR Range}  &  \textbf{0.6-1.0} & \textbf{1.2-2.0} & \textbf{2.2-2.6}  \\
\hline
Our $O(N^{1.5})$ detector & \textbf{0.47} & \textbf{0.75} & \textbf{0.88} \\
\hline
Our $O(N\log N)$ detector & \textbf{0.46} & \textbf{0.72} & \textbf{0.87} \\
\hline
Canny   & 0.24 & 0.67 & 0.85 \\
\hline
BM3D+Canny  & 0.12 & 0.67 & 0.84 \\
\hline
Crisp  & 0.3 & 0.32 & 0.46 \\
\hline
Dollar & 0.18 & 0.32 & 0.43 \\
\hline
MCG & 0.25 & 0.28 & 0.41 \\
\hline
\end{tabular}
\end{center}
\caption{Average F-measures obtained in simulations. The score is computed over images in three SNR ranges: from $0.6$ to $1$, $1.2$ to $2$ and $2.2$ to $2.6$.}
\label{tab:F-measure}
\end{table}

\begin{figure}
\centering

\fbox{\includegraphics[height=2cm]{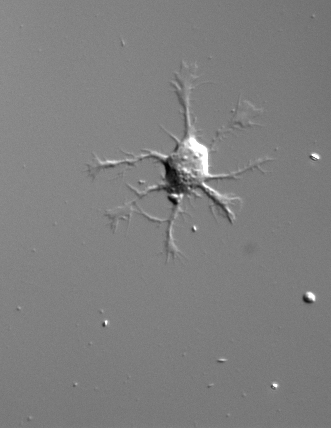}}~
\fbox{\includegraphics[height=2cm]{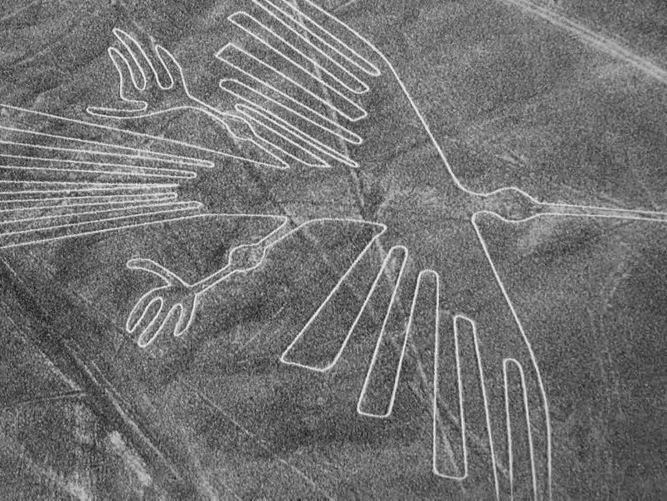}}~
\fbox{\includegraphics[height=2cm]{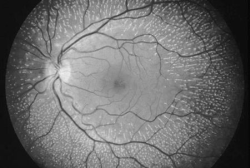}}
\\[0.1cm]
\fbox{\includegraphics[height=2cm]{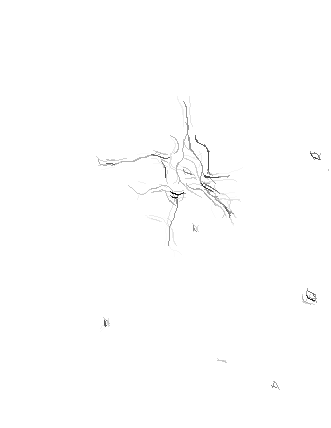}}~
\fbox{\includegraphics[height=2cm]{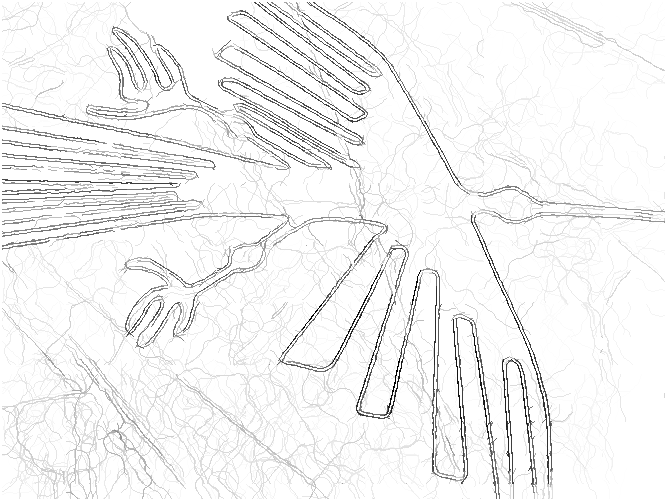}}~
\fbox{\includegraphics[height=2cm]{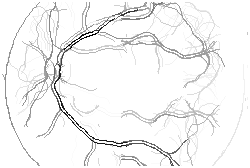}}
\\[0.1cm]
\fbox{\includegraphics[height=2cm]{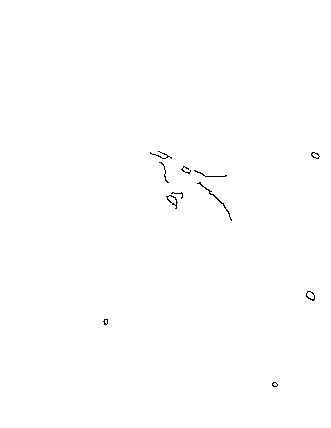}}~
\fbox{\includegraphics[height=2cm]{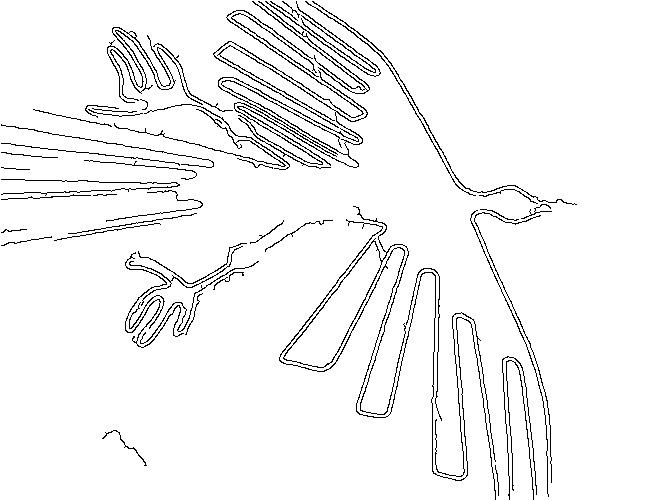}}~
\fbox{\includegraphics[height=2cm]{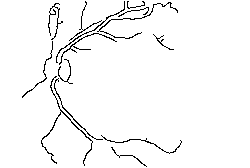}}
\\[0.1cm]
\fbox{\includegraphics[height=2cm]{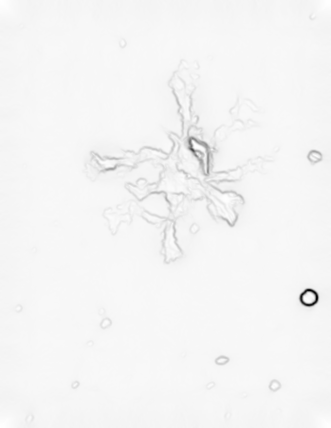}}~
\fbox{\includegraphics[height=2cm]{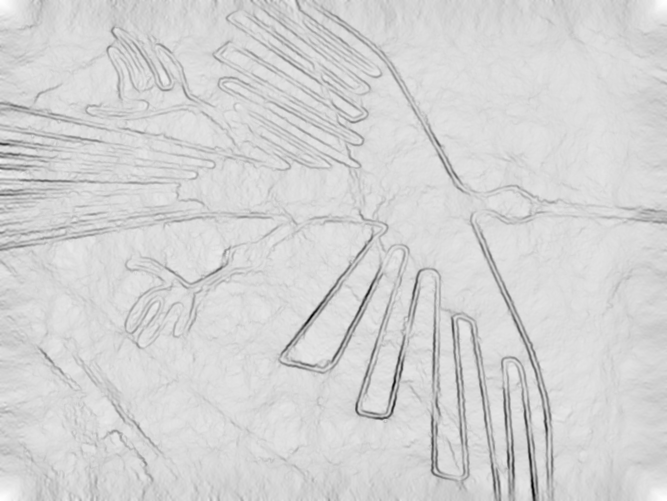}}~
\fbox{\includegraphics[height=2cm]{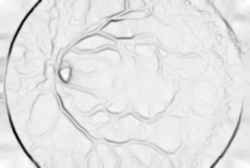}}
\caption{Real images: Each column shows the original image
and a comparison of our results with those obtained by other algorithms. From top to bottom: The input image; our $O(N^{1.5})$ algorithm; Canny and Crisp.}
\label{fig:Medical}
\end{figure}

\begin{figure}
\begin{center}
\fbox{\includegraphics[width=2.2cm]{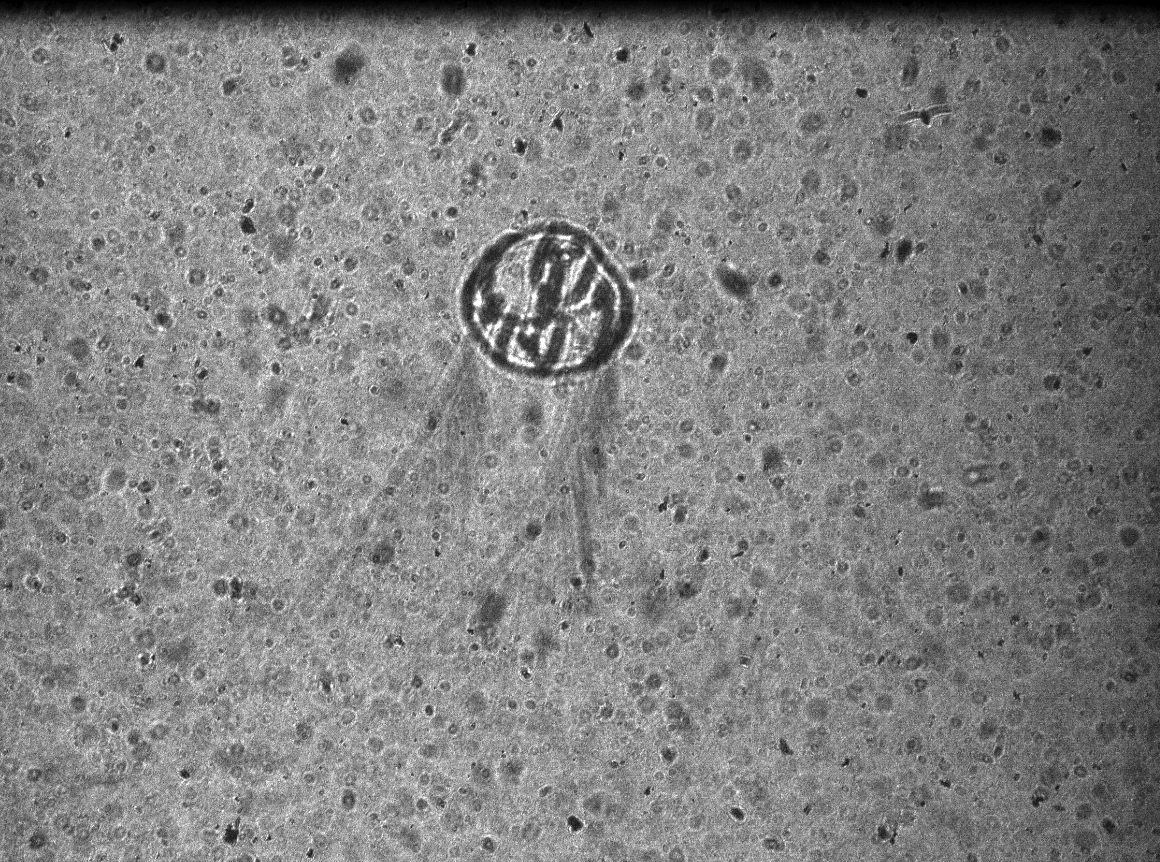}}~
\fbox{\includegraphics[width=2.2cm]{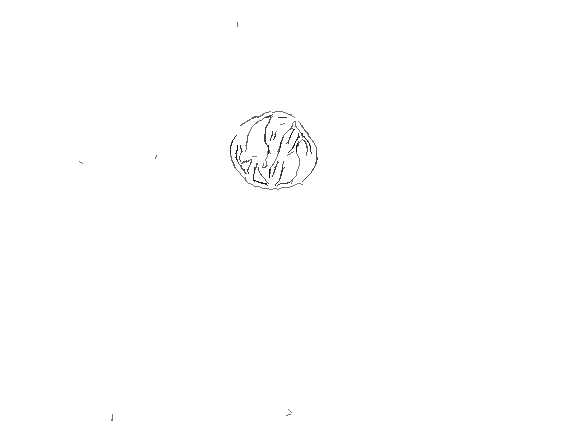}}~
\fbox{\includegraphics[width=2.2cm]{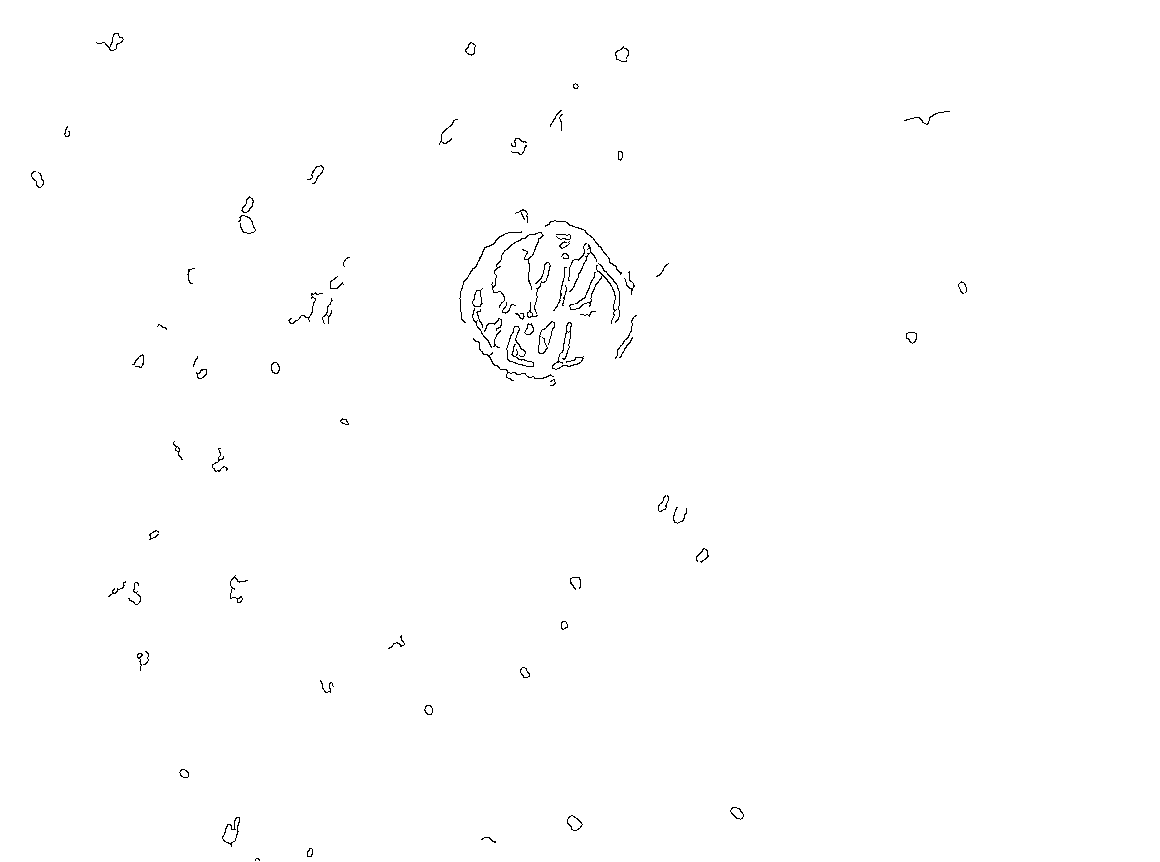}}
\\[0.1cm]
\fbox{\includegraphics[width=2.2cm]{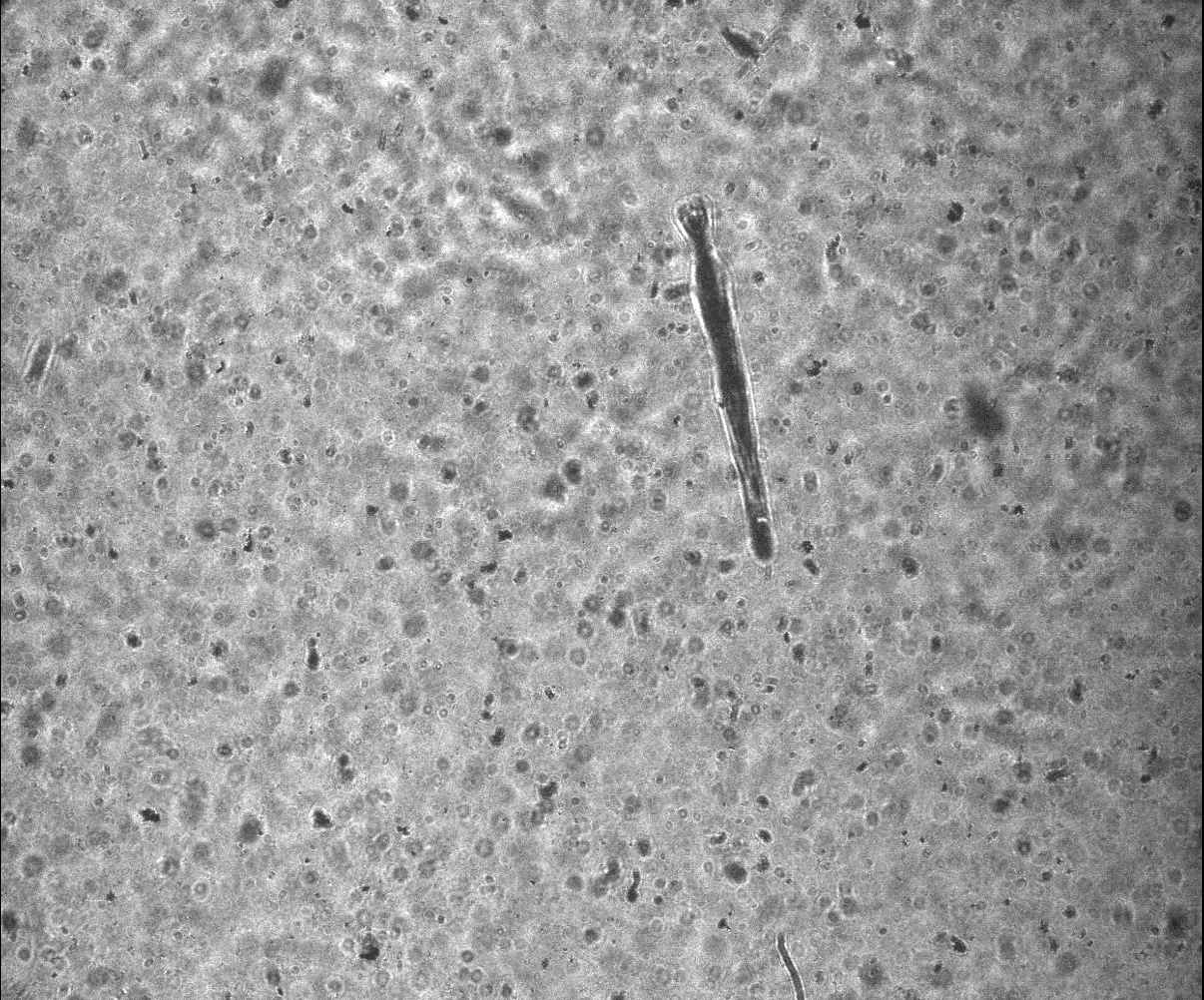}}~
\fbox{\includegraphics[width=2.2cm]{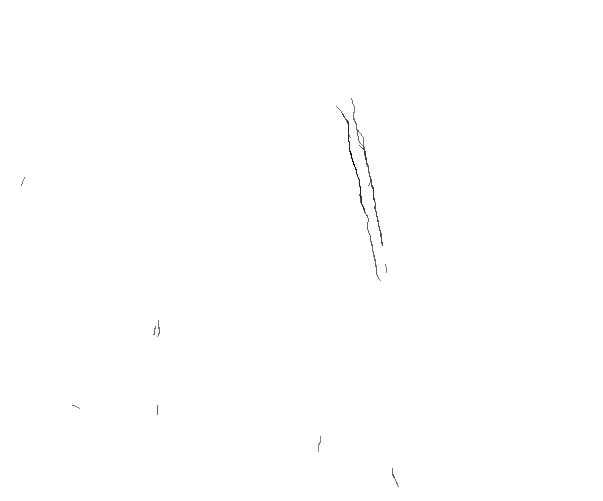}}~
\fbox{\includegraphics[width=2.2cm]{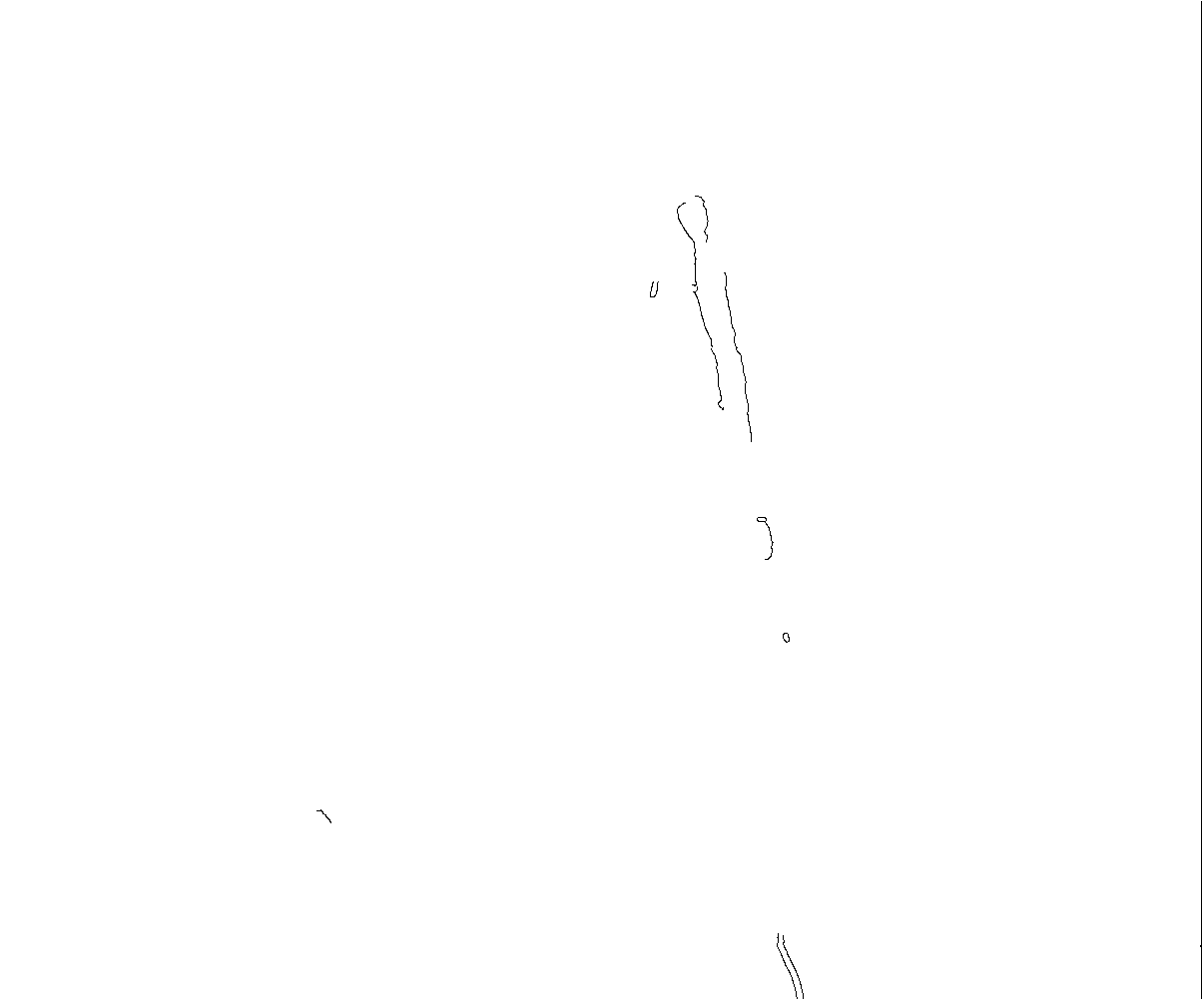}}
\end{center}
\caption{Underwater microscope example: Each row shows a Plankton image along with corresponding results. From left to right: the input image; our result and Canny result.}
\label{fig:plankton}
\end{figure}

\vspace{0.1cm}
\textbf{Real Images.}
Finally, we tested our algorithm on various real images taken under poor conditions. Fig.~\ref{fig:Medical} shows results on few such images. It can be seen that our method manages to accurately detect edges, even in challenging images where other approaches exhibit poor results. This is yet another empirical demonstration that state-of-the-art algorithms for boundary detection~\cite{Crisp,Dollar,MCG} may perform poorly in detecting edges at low SNRs. Canny, while significantly better, still fails to detect very faint and thin edges. Fig.~\ref{fig:plankton} shows yet another example of our and Canny results on underwater images of Plankton~\cite{Plankton}.


\section{Conclusion}

We presented efficient algorithms for detecting faint curved edges in noisy images that achieve state-of-the-art results in low SNRs. We introduced a novel approach for detecting curved edges in an image in $O(N\log N)$ operations. Importantly, the algorithm is adaptive to various parameters such as edge length, shape, and SNR. Thus it may be applicable in a variety of imaging domains.
Our approach can be extended to efficiently detect edges in 3D images, which are frequently used in medical imaging. In addition, it may be used to find maximum paths in other detection tasks such as visual object tracking. Source code is available on line.

{\small
\bibliographystyle{ieee}
\bibliography{bib}
}
\end{document}